\documentclass[a4paper,10pt]{article}
\usepackage[english]{babel}
\usepackage[utf8]{inputenc}

\usepackage{arxiv}

\usepackage[utf8]{inputenc} % allow utf-8 input
\usepackage[T1]{fontenc}    % use 8-bit T1 fonts
\usepackage{hyperref}       % hyperlinks
\usepackage{url}            % simple URL typesetting
\usepackage{booktabs}       % professional-quality tables
\usepackage{amsfonts}       % blackboard math symbols
\usepackage{nicefrac}       % compact symbols for 1/2, etc.
\usepackage{microtype}      % microtypography
\usepackage{lipsum}        % Can be removed after putting your text content

\usepackage{graphicx}   % for image

\usepackage{float}  % for table
\restylefloat{table}    % for table

\usepackage{amsmath} % for case equation

\usepackage{changepage}
\usepackage{color}

\title{\emph{Virtual Piano using Computer Vision} }

\date{}                     % Or removing it

\author{
  Seongjae~Kang,~Jaeyoon~Kim,~Sung-eui~Yoon\\
  KAIST, Daejeon, Republic of Korea \\
  \texttt{tjdwo2744@kaist.ac.kr,~jaeyoon1603@gmail.com,~sungeui@kaist.ac.kr} \\
}

\renewcommand{\headeright}{}

\begin{document}
\maketitle

\begin{abstract}
% reason to solve the problem
 In this research, Piano performances have been analyzed only based on visual information. Computer vision algorithms, e.g., Hough transform and binary thresholding, have been applied to find where the keyboard and specific keys are located.
 At the same time, Convolutional Neural Networks(CNNs) has been also utilized to find whether specific keys are pressed or not, and how much intensity the keys are pressed only based on visual information. Especially for detecting intensity, a new method of utilizing spatial, temporal CNNs model is devised. 
%  new method I used HERE :)
%  that spatial CNNs model and spatial, temporal CNNs model are trained for each task. 
 Early fusion technique is especially applied in temporal CNNs architecture to analyze hand movement. We also make a new dataset for training each model. Especially when finding an intensity of a pressed key, both of video frames and their optical flow images are used to train models to find effectiveness.
\end{abstract}

\section{Introduction}

 With the development of machine learning and deep learning, there have been many achievements in various fields. Especially computer vision field has been deeply affected by this technology like CNN and there have been achievements, for example, in image recognition, classification and so on \cite{krizhevsky2012imagenet, simonyan2014very, he2016deep}. Furthermore, there have been analyses of more than a static image. Especially, Video recognition, which is an important part of this research, has been done by many researchers. Unlike analyzing static images, analyzing video is much more complex by nature since it contains temporal information more than a spatial one. And CNN based video analysis shows it can be useful to analyze videos \cite{karpathy2014large}. And for the motion detection problem, it is well known that optical flow based analysis can be utilized for better precision \cite{simonyan2014two, carreira2017quo}. 
 \newline
 
 People get inspiration from music when it is consumed with visual factors not only acoustic factors. That is probably why people like to enjoy music with a live performance or music videos. More than that, the visual factor of music contains lots of information about the music itself. For example, one can notice which note or how strong music performer plays the music by watching the player’s hand movement and slight changes in the instrument itself. With this fact, there have been many types of research that tried to analyze visual factors in music performance too. For example, there were researches that analyzed guitar performance using oscillation of guitar strings or researches that analyzed piano performance using pianist’s hands and key movements \cite{goldstein2018guitar, goodwin2013key, vishal2017paper, suteparuk2014detection, akbari2015real, akbari2018real}. However, previous researches on piano analysis have limitations that it can only detect which keys are on or off and how long the keys persist \cite{vishal2017paper, suteparuk2014detection, akbari2015real}. Considering that music performers express their emotion using various intensity of notes, it is important to know how strong the performer plays the music more than which notes or how long the notes are played.  
 \newline
 
 This paper is organized as follows. In Section 2, related works done by researchers will be introduced. In Section 3, algorithms applied to detect piano area and each key, and way to extract feature vectors for CNN models will be covered. Then actual training and validation process will be on Section 4 and conclusion for this research is dealt with in Section 5. The overall process diagram is illustrated in Figure 1.
 \newline
 
\begin{figure} [h]
  \centering
  \includegraphics[width=0.9\textwidth]{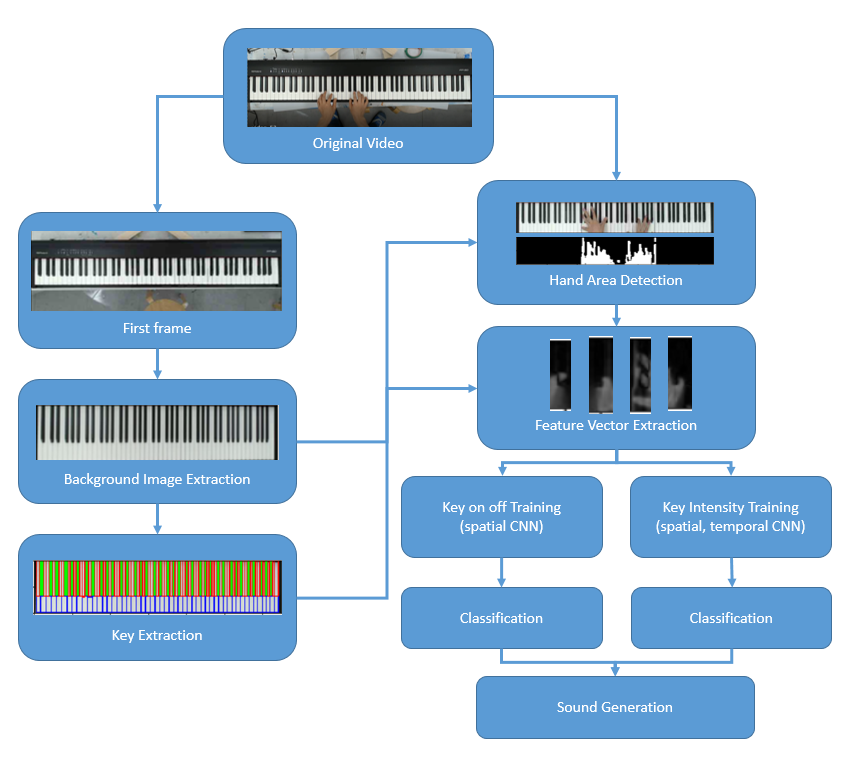}
  \caption{Overall diagram of the research}
  \label{fig:fig1}
\end{figure}

\section{Related Works}

 In this section, we first review previous studies on visual piano music and temporal analysis of videos. Loss functions for unbalances dataset and dataset with continuous labels are then discussed.

\subsection{Piano music analysis}

\newblock In order to analyze piano music based on visual information, the piano keyboard and each key should be detected to do further process. From the fact that piano keyboard has a rectangular shape and that each piano key is either white or black, Goodwin and Green demonstrated that the piano keyboard and each key can be detected using computer vision algorithms like Hough Transformation and binary thresholding \cite{goodwin2013key}. 
 
 For analyzing piano performance, various approaches were proposed \cite{vishal2017paper, suteparuk2014detection, akbari2015real, akbari2018real}. Vishal and Lawrence proposed that fingers can be tracked using shadow made by an artificial light directing on the keyboard \cite{vishal2017paper}. It is also shown that that the difference image between the current video frame and the background image can be utilized to detect piano key intensity change \cite{suteparuk2014detection}. Akbari and Cheng developed this idea with the fact that edge intensity changes when keys are press, and showed high accuracy under a certain condition \cite{akbari2015real}. Recently, they also analyzed the piano performance using a Convolutional Neural Networks (CNNs) architecture, and it had a high accuracy even under non-ideal camera views \cite{akbari2018real}.
 
\subsection{CNN-based temporal analysis on video}

 Videos contain not only spatial information but also temporal information, and that is why it is more challenging to analyze video than just an image. For CNN-based approaches, various image fusing methods were discussed according to the purpose of architectures \cite{karpathy2014large}. By fusing images in a certain stage of an architecture, it is shown that temporal information can be effectively analyzed depending on the purpose. For kinetics dataset, Carreira and Zisserman compared various approaches on CNN architecture designs \cite{carreira2017quo}. Various methods like LSTM, 3D-ConvNet, two-stream 3D-ConvNet were discussed especially on classifying kinetics dataset. Among these approaches, two-stream CNNs utilizing optical flow showed the highest accuracy on detecting human motion \cite{simonyan2014two, carreira2017quo}. 

\subsection{Loss function}

 For the classification task, the dataset could be inevitably unbalanced for certain reasons. In this case, just using whole data can lead to a biased result. This problem can be solved by changing loss function in a training model, by giving more weight to data labels which are rarer than other labels, which is descriped as focal loss in \cite{lin2017focal}.
 
 There is also an issue that labels are related to each other. Especially in age estimation using face images, a label distribution method was applied to make predicted distribution follow the Gaussian distribution \cite{gao2018age}. By adding Kullback–Leibler divergence loss with the original cross-entropy loss function, it took advantage of classifying the age of a person based on the fact that people with similar ages have a high chance to be detected as the same age.

\section{Detecting Piano Keyboard}

 In this section, methods to detect keyboard area and to extract feature vectors for CNN models will be addressed. 
More on that, the background image will be extracted in this process and will be used to make a feature vector for  future use. Additionally, there will be a hand area map created for efficiency. 

\subsection{Detecting Piano Keyboard}

 In this part, basic computer vision algorithms to find piano keyboard area and each piano key are explained in a given original image (Figure 2).
 
\begin{figure} [h]
  \centering
  \includegraphics[width=.7\textwidth]{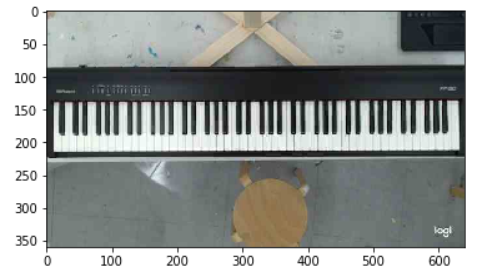}
  \caption{Original input image to detect keyboard}
  \label{fig:fig1}
\end{figure} 

 From the given original image, the first step is to find the main borderlines that form the piano keyboard area, which is rectangular seen in Figure 2. For this, the Hough Transformation algorithm is utilized to find the main lines in the original image. After this process, there remain candidate lines for  representing surrounding piano keys. So among each combination of lines, we find a proper rectangular, in which upper part of the area has many low-intensity components (e.g., black keys), and lower part of the area has many high-intensity components (e.g., white keys). Figure 3 shows the result of Hough transformation, and Figure 4 is a result of this process.

\begin{figure} [h]
  \centering
  \includegraphics[width=.6\textwidth]{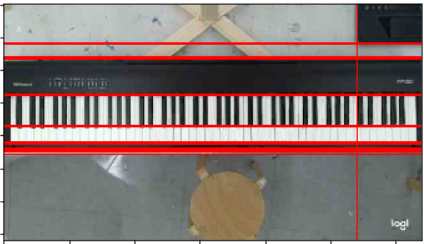}
  \caption{Result of Hough transformation on the original image}
  \label{fig:fig1}
\end{figure} 

\begin{figure} [h]
  \centering
  \includegraphics[width=.6\textwidth]{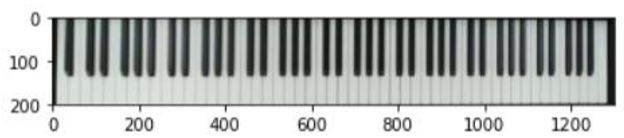}
  \caption{Result of finding keyboard area, which serve as our background image.}
  \label{fig:fig1}
\end{figure} 

 From this step, the resulting image will be used as a background image for the rest of the whole processes. Each key is then extracted from this background image as follows. First, we apply adaptive thresholding to distinguish white and black keys. By doing so, black keys can be found directly. For the white keys, the symmetry of the standard piano keyboard layout is used to find white keys. As a result, all keys can be extracted as follows (Figure 5).

\begin{figure} [h]
  \centering
  \includegraphics[width=.6\textwidth]{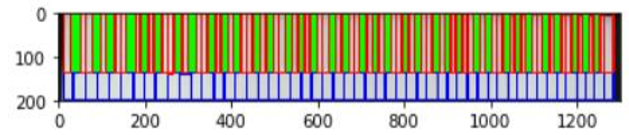}
  \caption{Extracted white and black keys}
  \label{fig:fig1}
\end{figure} 

 In Figure 5 above, the green squares represent extracted black keys, and red, blue squares represent extracted upper white keys and lower white keys, respectively.

\subsection{Hand Area Detection}

 Since the purpose of this research is to make virtual piano where one can play the piano in real-time, it is important to make an efficient algorithm. For this reason, the only interesting region will be used in training and predicting process.
This region is called hand area which is the area overlapped with hand and piano keyboard in the input image. As a result, a hand area detection algorithm is devised to make it efficient.

\begin{figure} [h]
  \centering
  \includegraphics[width=.6\textwidth]{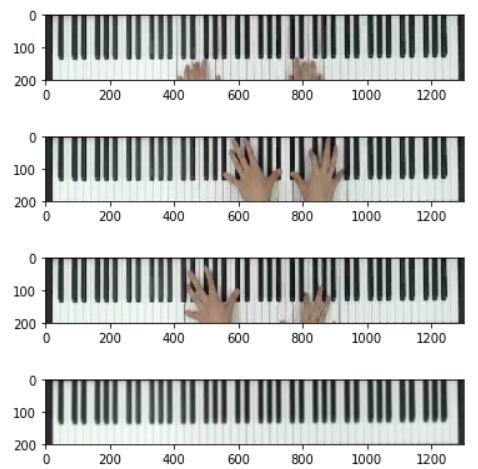}
  \caption{Original sample images and background image}
  \label{fig:fig1}
\end{figure}

 First, three images in Figure 6 were taken as samples for explaining hand area detection algorithm, and the last image is the background image extracted in the previous step. For detecting hand area, we get the difference image between sample and background images. After that, a noise reduction algorithm is applied to adjust slight camera move or light change.
 Figure 7 shows results after this process.
 White regions in Figure 7 are the detected hand region. From this, X coordinates where the regions white are located are saved. And only the pixels with respect to saved X coordinates will be processed in extracting feature vectors, training the models, and predicting.

 \begin{figure} [h]
  \centering
  \includegraphics[width=.6\textwidth]{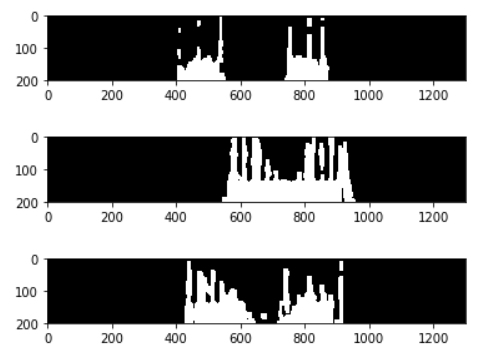}
  \caption{Detected hand region from Figure 6.}
  \label{fig:fig1}
\end{figure} 

\subsection{Extracting Feature Vectors}

 To apply machine learning algorithms, all keys should be transformed into proper training data set. For this, difference image, which is image with hand area subtracted by the background image, is first calculated. Then only for the keys where there are hand overlapped regions are considered as candidates for generating feature vector. From this difference image, the keys are transformed to proper size, and these transformed images are used as feature vectors.
 Specifically, each upper white keys and lower white keys are converted to 10 x 40 and 10 x 20 resolution images and combined into 10 x 60 resolution feature vector images. Black keys are converted directly into 10 x 40 resolution feature vector images.

 \begin{figure} [h]
  \centering
  \includegraphics[width=.6\textwidth]{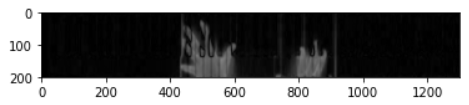}
  \caption{Sample difference image}
  \label{fig:fig1}
\end{figure} 

 \begin{figure} [h]
  \centering
  \includegraphics[width=.2\textwidth]{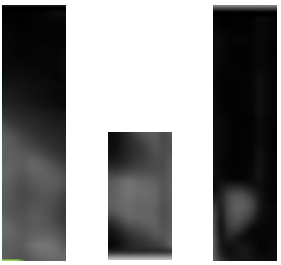}
  \caption{Examples of three extracted feature vectors from Figure 8}
  \label{fig:fig1}
\end{figure} 

 Figure 8 and 9 show examples of the difference image and extracted feature vectors. In Figure 9, first vector is 10 x 40 upper white key, second vector is lower 10 x 20 white key, and last one is 10 x 40 black key from Figure 9.

\section{Analysis on Piano Music using Convolutional Neural Network}

 In this section, we will discuss our training process.
 First, Dataset generation process will be explained as one part.
 And there will be explanation about process which detects which keys are pressed and how much intensity the keys are pressed in.
 As shown in Figure 1, detecting whether the keys are pressed or not and detecting how much intensity the keys are pressed go through different CNN models, since they need different information, each spatial only and both spatial, temporal information. 
 For this reason, this will be divided into two parts for each case. 

\subsection{Dataset}

\begin{figure} [h]
 \centering
 \includegraphics[width=.5\textwidth]{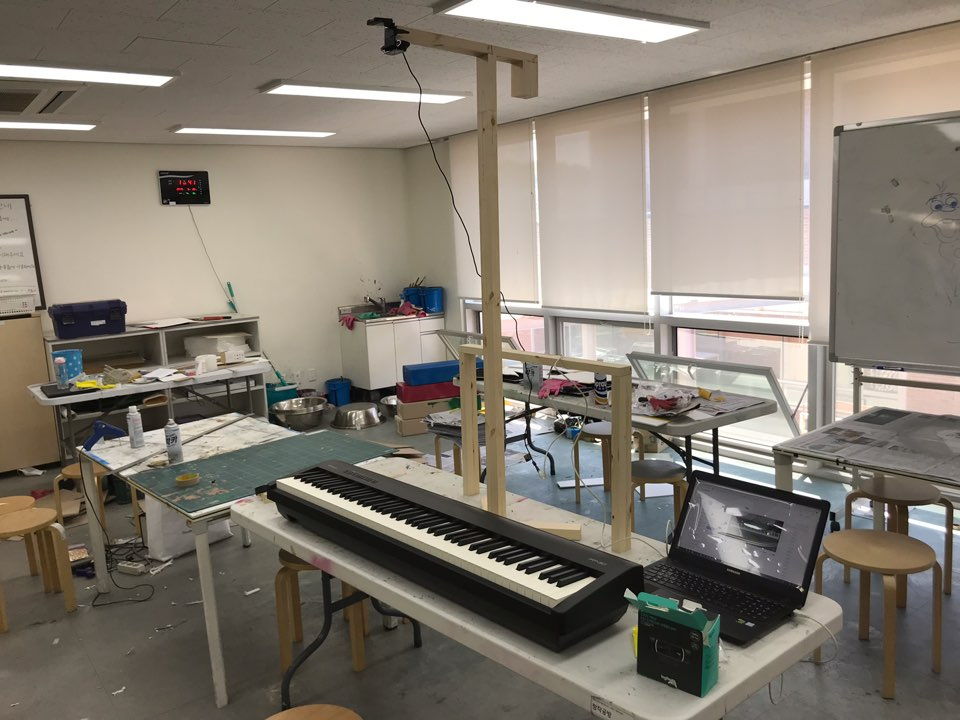}
 \caption{Testing environment}
 \label{fig:fig1}
\end{figure} 

\begin{figure}[htp]
 \centering
 \includegraphics[width=.33\textwidth]{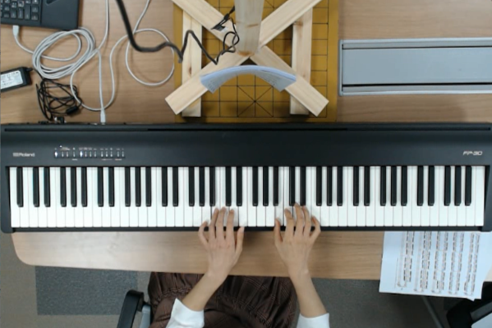}\hfill
 \includegraphics[width=.33\textwidth]{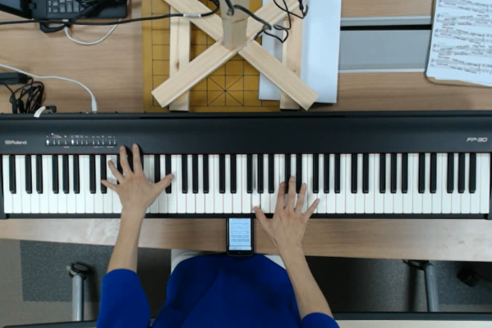}\hfill
 \includegraphics[width=.33\textwidth]{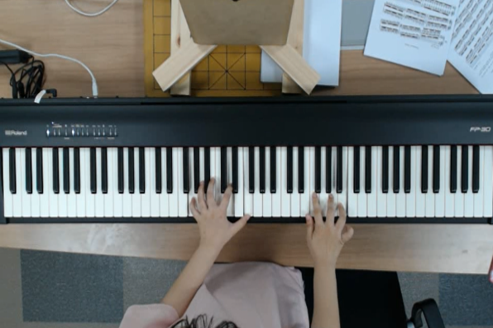}
 \caption{Example frames from sample videos}
 \label{fig:figure3}
\end{figure}
 
Since there was no dataset available from other sources, a new dataset is made manually in the testing environment (Figure 10). For detail, a Logitech c922 web camera is used to record silent video of the piano performance (30 frames per second, 960 x 640 resolution). A Roland FP 30 digital piano is used to send MIDI messages from the piano to computer using serial communication. Both silent video and MIDI messages are collected simultaneously in computer and combined into an original data. Dataset was made by a total of 10 random participants with different size of hands, playing styles, etc. Hanon score and real pieces of piano music were recorded by participants, and a total of 28 videos were taken as shown in Figure 11. Two kinds of our dataset are made as follows.
\newline

\begin{itemize}
\item Our first dataset is for detecting whether the key is pressed or not. Each training input consists of one feature vector image and its label, 0 when the key is not pressed and 1 when the key is pressed. For each white and black keys, total 6,755,380 and 4,680,813 image vectors were extracted from raw video data. Extracted data was highly unbalanced, that is, 6,261,509 white key data were extracted when keys are not pressed and the rest 493,871 were extracted when keys are pressed. Similarly, 4,500,158 black key data were extracted when keys are not pressed and the rest 180,655 data when keys are pressed.

\item Second dataset is for detecting how much intensity the key is pressed. Since training input should contain temporal information to detect hand movement of a piano player, feature vector images of the current frame, 2/15 seconds, 4/15 seconds, 6/15 seconds, 8/15 seconds before the current frame were combined as one training input. The intensity label was divided into five different levels to train a classification model, from 0 to 4 in increasing order of intensity. For each white and black keys, total 54,365 and 166,47 data were collected. The distribution of each data was from 1,601, 6,637, 21,705, 20,898, to 3,524 for white keys, and from 319, 1,305, 5,742, 8,216 to 1,065 for black keys, with increasing order of intensity. 

\item Last dataset is also for detecting how much intensity the key is pressed, but the dataset is made from optical flow maps. 6 feature vector images from the current frame, 2/15 seconds, 4/15 seconds, 6/15 seconds, 8/15 seconds, 10/15 seconds before the current frame were chosen, and optical flow maps between each feature vector images were calculated. These optical flow maps were then merged into training input for the training model. The number of the dataset is exactly the same as the previous dataset for detecting intensity since dataset was made from the same raw videos.
\end{itemize}

\subsection{Training for Detecting Keys to be On/Off}

 Detecting whether the keys are press or not requires just spatial information since changes in the border area of the keys are critical information to detect whether the keys are on or not. 
 
 Followings are architecture I used for this research (Table 1)
 \newline

\begin{table} [H]
 \begin{adjustwidth}{-0.8cm}{}
 \caption{Architecture for detecting whether keys are pressed or not}
  \centering
  \begin{tabular}{ll}
  \begin{tabular}{lll}
    \toprule
    Operation     & Kernel Size     & Output Size \\
    \midrule
    Input feature vector &   & (n, 600)     \\
    Reshape &   & (n, 10, 60, 1d)     \\
    Conv2D (Stride=1) & (3, 3, 1d, 16d) & (n, 10, 60, 16d)     \\
    ReLU &   & (n, 10, 60, 16d)     \\
    Max\_Pool (Drop\_Out) & (1, 2, 2, 1) & (n, 5, 30, 16d)     \\
    Conv2D (Stride=1) & (3, 3, 16d, 32d) & (n, 5, 30, 32d)     \\
    ReLU &   & (n, 5, 30, 32d)     \\
    Max\_Pool (Drop\_Out) & (1, 2, 2, 1) & (n, 3, 15, 32d)     \\
    Reshape &   & (n, 3 x 15 x 32)     \\
    Dense &   & (n, 256)     \\
    Dense &   & (n, 2)     \\
    \bottomrule
  \end{tabular}
  \begin{tabular}{lll}
    \toprule
    Operation     & Kernel Size     & Output Size \\
    \midrule
    Input feature vector &   & (n, 400)     \\
    Reshape &   & (n, 10, 40, 1d)     \\
    Conv2D (Stride=1) & (3, 3, 1d, 16d) & (n, 10,40, 16d)     \\
    ReLU &   & (n, 10, 40, 16d)     \\
    Max\_Pool (Drop\_Out) & (1, 2, 2, 1) & (n, 5, 20, 16d)     \\
    Conv2D (Stride=1) & (3, 3, 16d, 32d) & (n, 5, 20, 32d)     \\
    ReLU &   & (n, 5, 20, 32d)     \\
    Max\_Pool (Drop\_Out) & (1, 2, 2, 1) & (n, 3, 10, 32d)     \\
    Reshape &   & (n, 3 x 10 x 32)     \\
    Dense &   & (n, 256)     \\
    Dense &   & (n, 2)     \\
    \bottomrule
  \end{tabular}  
  \end{tabular}
  \label{tab:table}
  \end{adjustwidth}
\end{table}

 Training inputs are reshaped to the original feature vector image size, and go through 2D CNNs with ReLU, max pooling, and dropout layer two times. They pass fully connected layer two times with 256 and 2 output each. Since the original dataset is highly unbalanced in that data for the key not pressed exceed data for the key pressed with a large amount, focal loss function which gives more weight to data labels which are rarer than other label was utilized as a cost function \cite{lin2017focal}. Adam optimizer is used as an optimizer with varying learning rates slightly from 0.001 to 0.0001. Total epoch used in this experiment is 30. The result is shown in Table 2 below. As a result, it shows about 92.0\% and 94.1\% accuracy for detecting whether each white and black keys are pressed or not.

\begin{table} [H]
 \caption{Experimental results (Detecting whether keys are pressed or not)}
  \centering
  \begin{tabular}{ll}
    \toprule
    White or Black key     & Accuracy    \\
    \midrule
    White Key On Off & 0.919544    \\
    Black Key On Off & 0.941815    \\
    \bottomrule
  \end{tabular}
  \label{tab:table}
\end{table}

\subsection{Training for Detecting Intensity}

Detecting intensity requires analyzing temporal information since the velocity of hand movement affects the intensity of keys pressed. To deal with this problem, five feature vector images in time sequence are combined together to contain temporal information. For this research, an early  fusion method is utilized, since it has been known that it is useful to use early fusion structure to detect instantaneous motion by filtering temporal components at the beginning of the architecture \cite{karpathy2014large}. Since it is also known that early fusion has a simple structure, it is suitable for real-time purpose compared to other structures like later fusion. The whole architecture process is shown in Table 3 below. 
 
\begin{table} [H]
 \begin{adjustwidth}{-1.0cm}{}
 \caption{Architecture for detecting how much intensity keys are pressed with.}
  \centering
  \begin{tabular}{ll}
   \begin{tabular}{lll}
    \toprule
    Operation     & Kernel Size     & Output Size \\
    \midrule
    Input feature vector &   & (n, 3000)     \\
    Reshape &   & (n, 5, 10, 60, 1d)     \\
    Conv3D (Stride=1) & (5, 1, 1, 1d, 16d) & (n, 10, 60, 16d)     \\
    Conv2D (Stride=1) & (3, 3, 1d, 16d) & (n, 10, 60, 16d)     \\
    ReLU &   & (n, 10, 60, 16d)     \\
    Max\_Pool (Drop\_Out) & (1, 2, 2, 1) & (n, 5, 30, 16d)     \\
    Conv2D (Stride=1) & (3, 3, 16d, 32d) & (n, 5, 30, 32d)     \\
    ReLU &   & (n, 5, 30, 32d)     \\
    Max\_Pool (Drop\_Out) & (1, 2, 2, 1) & (n, 3, 15, 32d)     \\
    Reshape &   & (n, 3 x 15 x 32)     \\
    Dense &   & (n, 256)     \\
    Dense &   & (n, 5)     \\
    \bottomrule
    \end{tabular}
  \begin{tabular}{lll}
    \toprule
    Operation     & Kernel Size     & Output Size \\
    \midrule
    Input feature vector &   & (n, 2000)     \\
    Reshape &   & (n, 5, 10, 40, 1d)     \\
    Conv3D (Stride=1) & (5, 1, 1, 1d, 16d) & (n, 10, 40, 16d)     \\
    Conv2D (Stride=1) & (3, 3, 1d, 16d) & (n, 10, 40, 16d)     \\
    ReLU &   & (n, 10, 40, 16d)     \\
    Max\_Pool (Drop\_Out) & (1, 2, 2, 1) & (n, 5, 20, 16d)     \\
    Conv2D (Stride=1) & (3, 3, 16d, 32d) & (n, 5, 20, 32d)     \\
    ReLU &   & (n, 5, 20, 32d)     \\
    Max\_Pool (Drop\_Out) & (1, 2, 2, 1) & (n, 3, 10, 32d)     \\
    Reshape &   & (n, 3 x 10 x 32)     \\
    Dense &   & (n, 256)     \\
    Dense &   & (n, 5)     \\
    \bottomrule
    \end{tabular}
  \end{tabular}
  \label{tab:table}
  \end{adjustwidth}
\end{table}

Training inputs are reshaped to a 3D structure with the time component multiplied by the original feature vector image size. They then go through 3D CNN stride 1 to apply early fusion.
%method. 
After that, they are reshaped again to reduce the size from 3D to a 2D image vector, followed by passing exactly the same architecture used in detecting whether keys are pressed or not, except the final fully connected layer that gives five outputs to represent five different intensity levels. Distribution labeling is utilized as a cost function since intensity label is highly likely to be related with its neighbor labels than others (e.g., intensity label 1 is more likely to be related with label 2 than label 3). Adam optimizer is used as an optimizer with a fixed learning rate of 0.001. Total epoch used in this experiment is 15. 

Optical flow maps were also utilized to detect intensity. Similar to combining 5 different feature vector images, training input was made by combining 5 different optical flow maps as mentioned in the dataset section. Since the optical flow map has x and y gradient components, the training input size is doubled.

\begin{table} [H]
 \begin{adjustwidth}{-1.0cm}{}
 \caption{Architecture for detecting whether keys are pressed or not (optical flow)}
  \centering
  \begin{tabular}{ll}
   \begin{tabular}{lll}
    \toprule
    Operation     & Kernel Size     & Output Size \\
    \midrule
    Input feature vector &   & (n, 6000)     \\
    Reshape &   & (n, 5, 10, 120, 1d)     \\
    Conv3D (Stride=1) & (5, 1, 1, 1d, 16d) & (n, 10, 120, 16d)     \\
    Conv2D (Stride=1) & (3, 3, 1d, 16d) & (n, 10, 120, 16d)     \\
    ReLU &   & (n, 10, 120, 16d)     \\
    Max\_Pool (Drop\_Out) & (1, 2, 2, 1) & (n, 5, 60, 16d)     \\
    Conv2D (Stride=1) & (3, 3, 16d, 32d) & (n, 5, 60, 32d)     \\
    ReLU &   & (n, 5, 60, 32d)     \\
    Max\_Pool (Drop\_Out) & (1, 2, 2, 1) & (n, 3, 30, 32d)     \\
    Reshape &   & (n, 3 x 30 x 32)     \\
    Dense &   & (n, 256)     \\
    Dense &   & (n, 5)     \\
    \bottomrule
    \end{tabular}
  \begin{tabular}{lll}
    \toprule
    Operation     & Kernel Size     & Output Size \\
    \midrule
    Input feature vector &   & (n, 4000)     \\
    Reshape &   & (n, 5, 10, 80, 1d)     \\
    Conv3D (Stride=1) & (5, 1, 1, 1d, 16d) & (n, 10, 80, 16d)     \\
    Conv2D (Stride=1) & (3, 3, 1d, 16d) & (n, 10, 80, 16d)     \\
    ReLU &   & (n, 10, 80, 16d)     \\
    Max\_Pool (Drop\_Out) & (1, 2, 2, 1) & (n, 5, 40, 16d)     \\
    Conv2D (Stride=1) & (3, 3, 16d, 32d) & (n, 5, 40, 32d)     \\
    ReLU &   & (n, 5, 40, 32d)     \\
    Max\_Pool (Drop\_Out) & (1, 2, 2, 1) & (n, 3, 20, 32d)     \\
    Reshape &   & (n, 3 x 20 x 32)     \\
    Dense &   & (n, 256)     \\
    Dense &   & (n, 5)     \\
    \bottomrule
    \end{tabular}
  \end{tabular}
  \label{tab:table}
  \end{adjustwidth}
\end{table}

As shown in Table 4, the architecture used in optical flow analysis is exactly the same as using stack feature vectors, with different training input size.

\begin{table} [H]
 \caption{Experimental results (Detecting how much intensity keys are pressed or not)}
  \centering
  \begin{tabular}{ll}
    \toprule
    White or Black key     & Accuracy    \\
    \midrule
    White Key Intensity & 0.532279    \\
    Black Key Intensity & 0.577778    \\
    White Key Intensity (Optical Flow) & 0.501931    \\
    Black Key Intensity (Optical Flow) & 0.512913    \\
    \bottomrule
  \end{tabular}
  \label{tab:table}
\end{table}

 There were two experiments for detecting intensity of key pressed, using stacked different images and using optical flow maps as inputs. They went through the same architecture explained before. As a result, it shows 53.2\% and 57.8\% accuracy for detecting each white and black key intensities respectively, when using stacked difference image vectors as an input. On the other hand, it shows  50.2\% and 51.3\% accuracy for detecting white and black key intensities, when using optical flow maps. 
 
 The result shows that optical flow is not good for detecting intensity compared to using just stacked difference images, though previous researches show that even optical flow based algorithm proved to be better than using stacked images in motion detection. It seems that the optical flow method shows an inferior result since this research deals with small motion such as up and down movement of hand unlike other studies dealing with large gestures like swinging arms. Another reason might be due to image size. Since image size is at most 10 x 60 for the case of white keys, optical flow result might be biased due to the small size of the image vector. 

\section{Conclusion}
In this article, it is shown that piano performance can be effectively analyzed with only visual information. By capturing silent video above a mock-up device of piano keyboard,  various features can be detected without acoustic information, such as whether the keys are pressed or not, how long the keys are pressed, and how much intensity the performer plays the keys. 


\nocite{*}
\bibliographystyle{unsrt}
\bibliography{reference}

\end{document}